\documentclass[final]{cvpr}

\usepackage{times}
\usepackage{epsfig}
\usepackage{graphicx}
\usepackage{amsmath}
\usepackage{amssymb}
\usepackage[normalem]{ulem}
\usepackage{CJKutf8}
\usepackage[utf8]{inputenc}
\usepackage{multirow}
\usepackage{booktabs}
\usepackage{tabularx}

\usepackage{color}
\definecolor{citecolor}{RGB}{119,185,0} 
\usepackage[pagebackref=true,breaklinks=true,letterpaper=true,colorlinks,citecolor=citecolor,bookmarks=false]{hyperref}

\def\cvprPaperID{8} 
\def\confYear{CVPR 2021}

\begin{document}

\title{Connecting Language and Vision \\ for Natural Language-Based Vehicle Retrieval}

\author{Shuai Bai\textsuperscript{1}  \quad Zhedong Zheng\textsuperscript{2}  \quad Xiaohan Wang\textsuperscript{3} \quad Junyang Lin\textsuperscript{1} \\ 
 Zhu Zhang\textsuperscript{1} \quad Chang Zhou\textsuperscript{1}   \quad Yi Yang\textsuperscript{2}  \quad Hongxia Yang\textsuperscript{1}\\
\textsuperscript{1}DAMO Academy, Alibaba Group,  \textsuperscript{2}ReLER Lab, University of Technology Sydney,  \textsuperscript{3} Zhejiang University\\
}

\pagestyle{empty}
\maketitle
\thispagestyle{empty}

\begin{abstract}
Vehicle search is one basic task for the efficient traffic management in terms of the AI City. Most existing practices focus on the image-based vehicle matching, including vehicle re-identification and vehicle tracking. In this paper, we apply one new modality, \ie, the language description, to search the vehicle of interest and explore the potential of this task in the real-world scenario. The natural language-based vehicle search poses one new challenge of fine-grained understanding of both vision and language modalities. 
To connect language and vision, we propose to jointly train the state-of-the-art vision models with the transformer-based language model in an end-to-end manner. Except for the network structure design and the training strategy, several optimization objectives are also re-visited in this work. The qualitative and quantitative experiments verify the effectiveness of the proposed method. Our proposed method has achieved the \textbf{1st} place on the 5th AI City Challenge, yielding competitive performance \textbf{$18.69\%$} MRR accuracy on the private test set. We hope this work can pave the way for the future study on using language description effectively and efficiently for real-world vehicle retrieval systems. The code will be available at \url{https://github.com/ShuaiBai623/AIC2021-T5-CLV}.
\end{abstract}

\section{Introduction}
Vehicle retrieval usually meets a large-scale candidate pool due to the 24/7 records, which is an important part of the intelligent transportation system for AI City. Most existing vehicle retrieval systems are based on image-to-image matching, also known as vehicle re-identification (vehicle re-id)~\cite{zheng2020vehiclenet}. To retrieve the target vehicle tracklets, these methods require a vehicle image query, which is not always available in the real-world scenario~\cite{lin2017improving,Feng21CityFlowNL}. In this report, we leverage natural language descriptions to search vehicles. Compared to image queries, natural language descriptions are more user-friendly and easier to be obtained. Besides, it enables fuzzy vehicle search and provides more flexible applications.

A common approach to performing language-based vehicle retrieval is to embed the images and descriptions to shared feature space and then rank the vehicle images based on the cross-modal similarities. Most existing methods~\cite{dong2019dual} construct the visual encoder and text encoder with fixed backbones and only optimize several projection layers. 
In our solution, we train the cross-modal vehicle retrieval framework in an end-to-end manner. This design enables the powerful backbones to learn fine-grained vehicle attributes like vehicle types and directions. 
Inspired by the recent advances in cross-modal representation learning~\cite{radford2021learning,zheng2017dual}, we adopt the symmetric InfoNCE loss~\cite{oord2018representation} and instance loss~\cite{zheng2017dual} to jointly train the text encoder and visual encoder. 

For the visual encoder, we propose a two-stream architecture to provide complimentary local details (\eg color, type and size) and global information (\eg motion and environment). The motivation behind the design is that the natural language sentences not only contain the information of vehicle appearance but also describe the trajectories and background. This is also the main difference between language-based vehicle retrieval and image-based vehicle retrieval.
Specifically, we adopt two individual CNNs to construct the backbones of the two streams. The local stream takes the detected patches that only contain the vehicle as input, while the input for the global stream is the synthesized dynamic image with the averaged background and the trajectory of the vehicle. The outputs from the two streams are concatenated as the final visual representation.
For the text encoder, we adopt the state-of-the-art transformer-based language models like BERT~\cite{devlin2018bert} and RoBERTa~\cite{liu2019roberta} as our text encoder. To enhance the robustness of the model, we propose a text augmentation approach by back-translation technique. The proposed method has achieved $18.69\%$ MRR accuracy on the private test set of the 5th AI City Challenge on natural language-based vehicle retrieval, yielding the \textbf{1st} place on the public leaderboard.



\section{Related Work}
\subsection{Video Retrieval via Natural Language}
Natural language-based video retrieval aims to search a specific video matching  the given language description from a large amount of candidate videos. Most existing works~\cite{lin2014visual,xu2015jointly,otani2016learning,yu2016video,yu2017end,mithun2018learning,zhang2018cross,miech2018learning,dong2019dual,wang2021} adopt the similarity learning~\cite{xing2002distance} to learn a function~(network) that can estimate the similarity between videos and language descriptions. 
These works encode the language by the textual feature extractor (Word2Vec~\cite{mikolov2013efficient}, LSTM~\cite{hochreiter1997long}, etc.), learn video representations by the visual feature extractor (Two-Stream Network~\cite{simonyan2014two}, C3D~\cite{carreira2017quo}, S3D~\cite{xie2018rethinking},~etc.) and estimate the language-video similarity in a common semantic space. 
Further, for language representations, Xu \etal~\cite{xu2015jointly} design a compositional language model by the dependency-tree structure. Yu \etal~\cite{yu2017end} develop a high-level concept detector as semantic priors and apply the attention mechanism to selectively focuses on the detected word concepts. 
For video representations, recent methods~\cite{sun2019learning,bain2021frozen} utilize the well-designed Transformer architecture~\cite{vaswani2017attention} to learn powerful video features. And some works~\cite{mithun2018learning,miech2018learning,gabeur2020multi,dzabraev2021mdmmt} further incorporate multi-modal features (\eg motion, audio) from a video for more robust video understanding. 
As for the video-language interaction, Zhang \etal~\cite{zhang2018cross} exploit both low-level and high-level correspondences in the hierarchically semantic spaces, and Dong \etal~\cite{dong2019dual} propose the multi-level encoding including global, local and temporal patterns in both videos and sentences to learn better shared representations.
Besides, M6~\cite{lin2021m6}, VideoBERT~\cite{sun2019videobert}, UniViLM~\cite{luo2020univilm}, HERO~\cite{li2020hero} and ClipBERT~\cite{lei2021less} explore the large-scale vision-language pre-training to boost comprehensive video-language understanding.

Recently, to search fine-grained video contents via natural language, researchers begin to explore moment retrieval and object retrieval in videos. Video moment retrieval~\cite{gao2017tall,hendricks2017localizing,zhang2020learning} localizes a video clip corresponding to the given language, which avoids manually searching for the clip of interests in a long video. Existing approaches often pre-define a series of clip proposals by  sliding windows or multi-granularity anchors, and rank these clips by visual-textual interaction and estimation, \eg attention mechanism~\cite{liu2018attentive} and graph convolution~\cite{zhang2019man}.
And video object retrieval~\cite{yamaguchi2017spatio,chen2019weakly,zhang2020does} aims to search the spatio-temporal object track (i.e. a sequence of bounding boxes) according to the language description. Early work~\cite{yamaguchi2017spatio} only searches the person track in multiple videos and recent approaches~\cite{zhou2018weakly,chen2019weakly} further retrieve the spatio-temporal tracks of diverse objects. 
Besides single-object retrieval, Huang and Shi \etal~\cite{Huang_2018_CVPR,shi2019not} try to localize multiple objects that appear in the language description.
Different from previous tasks, vehicle retrieval via natural language is one  practical task for the traffic management, which retrieves the specific vehicle given single-camera tracks and corresponding language descriptions of the targets. Our method sufficiently considers the inherent attributes of the vehicles as well as global motion and environment information to search the described vehicle.

\begin{figure*}[t]
\begin{center}
\includegraphics[width= 17cm]{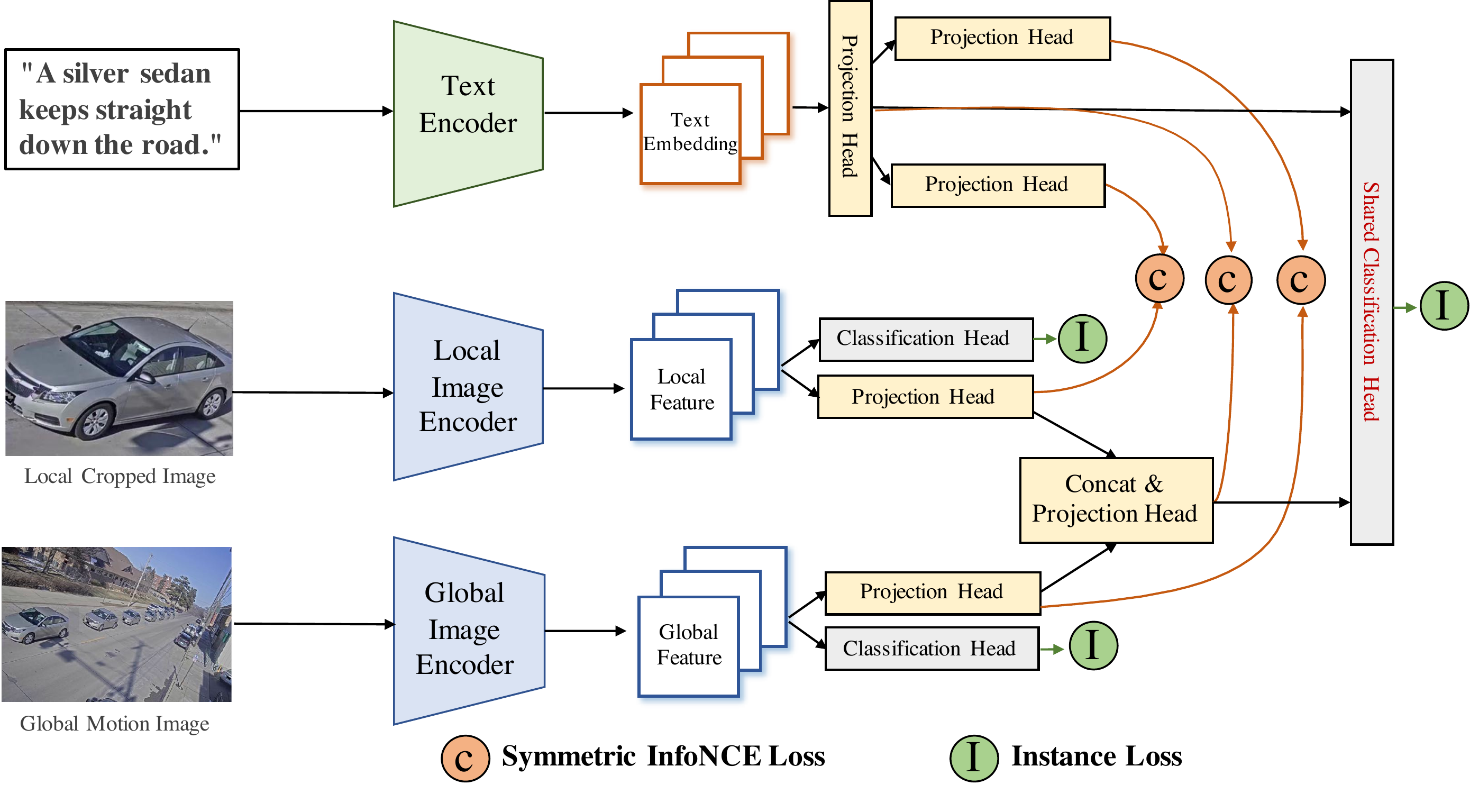}
   \caption{\textbf{The overall framework.}  The vanilla network only contains one local image encoder for local cropped vehicle image and one text encoder for language description input. We further introduce the global image encoder to help learning more position information as well as the environment from motion images to ease the matching difficulty. In this paper, we also explore different optimization objectives, including Symmetric InfoNCE Loss~\cite{oord2018representation} and Instance Loss~\cite{zheng2017dual}. }
   \label{fig:framework}
\end{center}
\end{figure*}

\subsection{Vehicle Re-identification}
Vehicle re-identification (vehicle re-id) is to find the vehicle of interest from millions of candidate images from different cameras, which can largely save the human resources as well as the time cost. Several pioneering works focus on building the large-scale dataset for subsequential learning, including VehicleID~\cite{liu2016pku}, VeRi-776~\cite{liu2016deep} and VehicleNet~\cite{zheng2020vehiclenet}. The follow-up works focus on the discriminative representation learning ~\cite{liu2016deep,zheng2019joint} as well as mining the structure information~\cite{wang2017orientation,sun2017beyond}. For instance, Qian~\etal~\cite{qian2017multi} and Yu~\etal~\cite{yu2017devil} propose to leverage the multi-scale information within deeply-learned models. To mine the fine-grained vehicle structure, Wang~\etal~\cite{wang2017orientation} further take the keypoints into consideration and apply the structure information to the final feature aggregation part. Besides, Zheng~\etal~\cite{zheng2020vehiclenet} apply the transfer learning to distill common knowledge from large-scale vehicle dataset to the specific small dataset, yielding the state-of-the-art performance. Attributes and environment conditions also have been explored in several pioneering works~\cite{lou2019veri,lin2017improving}. 
In summary, vehicle re-identification is primarily different from the natural language-based vehicle retrieval in terms of the input modality. The two different modalities are inherently different, which is challenging in mapping heterogeneous inputs to the same semantic space. In this paper, we mainly focus on the vehicle tracklet retrieval via the natural language, but the existing vehicle re-id also gives us many inspirations in the representation learning and optimization strategies. We will provide more details in Section ~\ref{sec:method}.

\subsection{Data Augmentation in NLP}
Data augmentation has gradually become a common practice in NLP and it brings substantial improvement due to the requirement of a large amount of training data. 
The augmented data should be semantic-consistent variants of the original ones. 
A conventional method is lexical replacement, including synonym replacement with WordNet~\cite{zhang2015charconv, mueller2016siamese, wei2019eda}, word embedding substitution~\cite{jiao2020tinybert,wang2015annoying}, and masked language modeling~\cite{garg2020bae}, etc. 
Backtranslation is an effective method to generate samples that are semantically invariant~\cite{rico2015backtranslation}, and it strongly promotes the development of unsupervised machine translation~\cite{lample2018unsupervisedmt}. Xie~\etal~\cite{xie2020dataaug} applied backtranslation for text classification and reached the state-of-the-art performance. Other methods include random noise injection~\cite{xie2020dataaug, wei2019eda}, syntax tree manipulation~\cite{coulombe2018aug}, mixup~\cite{guo2019mixup}, etc. 

\section{Method}~\label{sec:method}
In this section, we provide one detailed illustration of the proposed framework. In particular, we first start with the data augmentation strategies, which include the motion modeling and description augmentation. Followed by the data augmentation, the representation learning contains the description of the network structure and optimization functions. 

\subsection{Data Augmentation}
\subsubsection{Motion and background modeling}

\begin{figure}[t]
\begin{center}
\includegraphics[width= 1\linewidth]{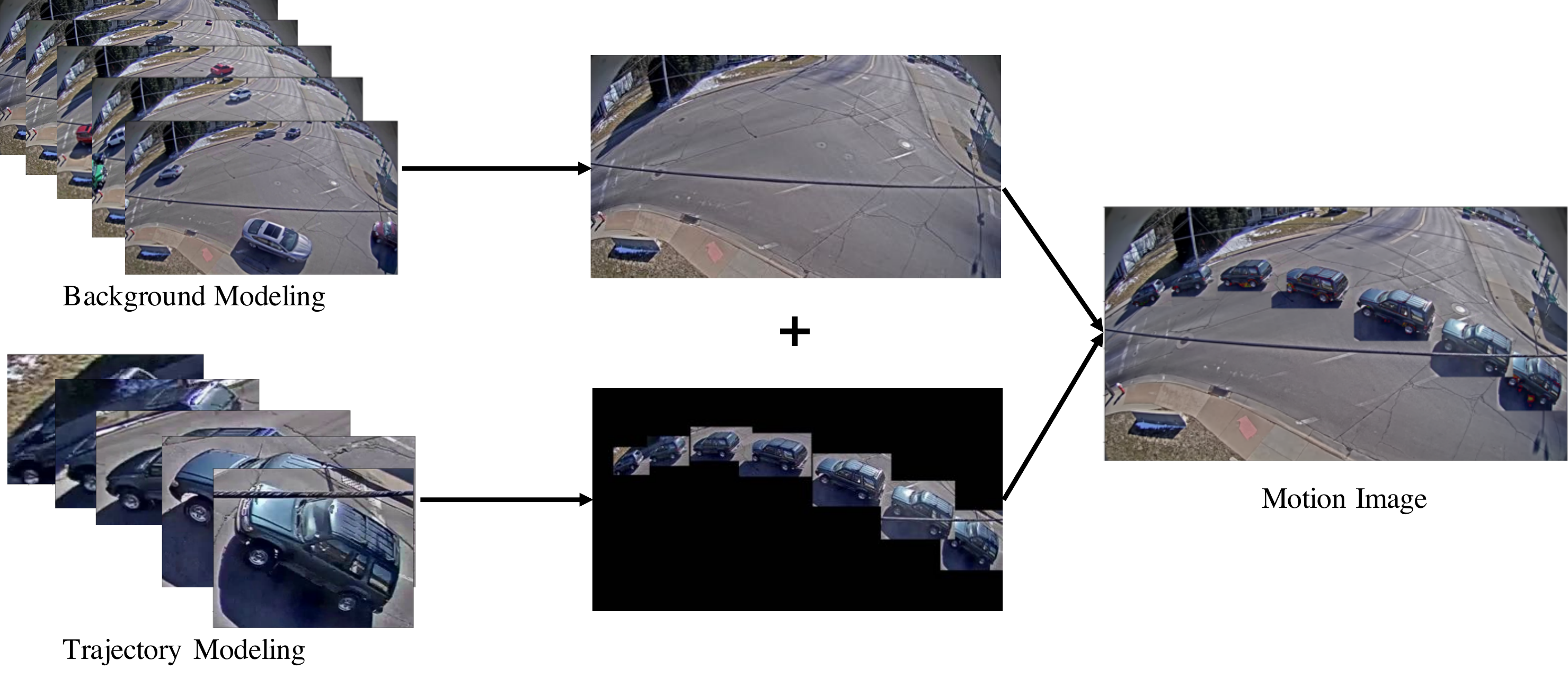}
   \caption{\textbf{Motion and background modeling.} Here we show the steps to obtain motion maps. In particular, we average the whole views into one consistent background, and then past the trajectory vehicle bounding boxes with the time gap. }
   \label{motion}
\end{center}
\end{figure}

Compared with the image-based vehicle retrieval, which applies vehicle images as queries for fine-grained appearance modeling, the natural language descriptions contain more surrounding factors and the motion information. The inherent attributes of vehicles are not enough to distinguish the specific target. For example, two white SUVs that go straight and turn left are difficult to distinguish only through the color and type of the vehicles. Therefore, the introduction of global information, such as background, is of vital importance for the accurate natural language-based vehicle retrieval. We propose a simple but effective way to introduce global information.  As shown in Figure~\ref{motion}, we adopt background and trajectory modeling to preserve environment and motion information as a motion image. In particular, the generation of motion images consists of three steps.

Firstly, we notice that the camera position is fixed, and maintains the same angle of view in video clips. It means that the background in the same video is stable. The method continuously calculating the weighted sum of input frame can enhance the static parts and remove the moving vehicles, which is widely applied in traffic anomaly detection \cite{bai2019traffic,xu2018dual}. Specifically, we calculate the mean value of each frame in the same video to generate background images. It can be formulated as: 
\begin{equation}
B = \frac{1}{N}\sum_i^N {F_i},
\end{equation}
where ${F_i}$ is the ${i_{th}}$ frame, ${B}$ is the background image, and ${N}$ is the number of video frames. For the AICity training/test data, the environmental information is preserved in the background image, including parking lots, intersections and traffic lights.

Secondly, the motion information of the vehicle reflects that the position of the vehicle is different at different moments. We continuously cover the crop of the detection box to the trajectory image $T$. In particular, since the movement distance of consecutive frames is limited, we use interval frames for coverage. 
\begin{equation}
T_{box_i} = F_{box_i}, 
\end{equation}
where $box_i$ is the detection box in the ${i_{th}}$ frame. $T$ is the trajectory image, and it is initialized with the zero matrix as large as the video frames.

Finally, we copy the detected bounding boxes, and paste the vehicles of different timestamps to  the background image as the motion image.

\subsubsection{Description Augmentation}
\begin{figure}[tb]
    \centering
    \includegraphics[width=0.45\textwidth]{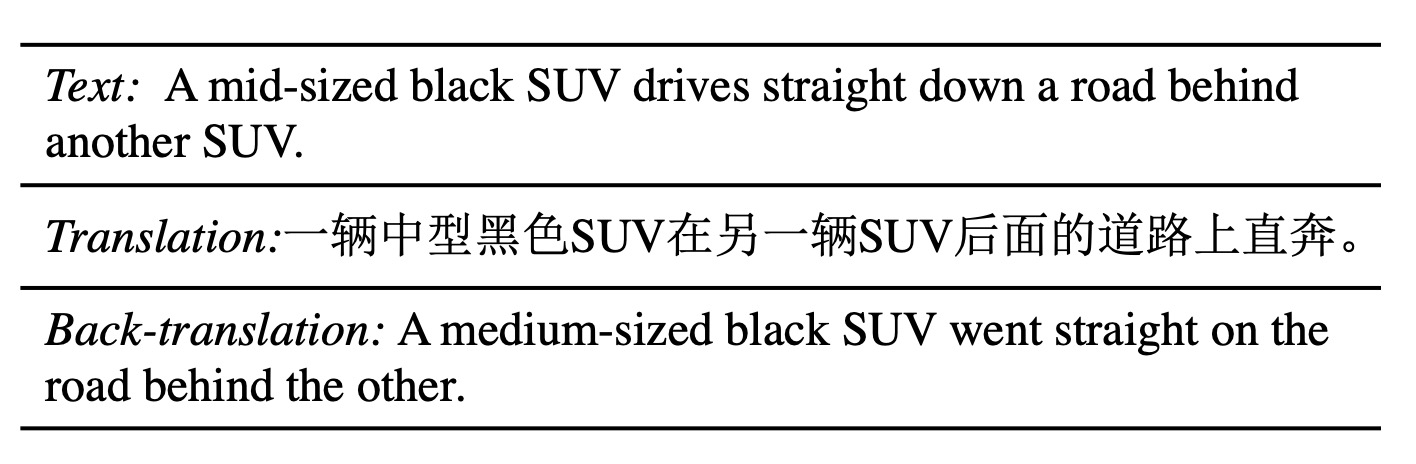}
    \caption{An example of the translation and back-translation as description augmentation. We first translate the original training sentence from English to Chinese, and then translate the sentence back to the English format.
    }
    \label{fig:aug_example}
\end{figure}

In order to provide more training data and enhance the model robustness, we apply text data augmentation. 
Specifically, in our practice, we use backtranslation to generate semantic invariants for the training samples. 
We collect all the command texts and apply translation and backtranslation with the in-house application Ali-translate. 
We observe the text data and find that the commands are mostly of short length and without complex syntactic structure. 
Translating to languages that are similar to English, such as French and German, may cause backtranslation's generating the same texts. 
Instead, we translate the texts to Chinese and backtranslate them to English. 
We demonstrate some examples of translation and backtranslation in Figure~\ref{fig:aug_example}.

\begin{figure}[tb]
    \centering
    \includegraphics[width=0.45\textwidth]{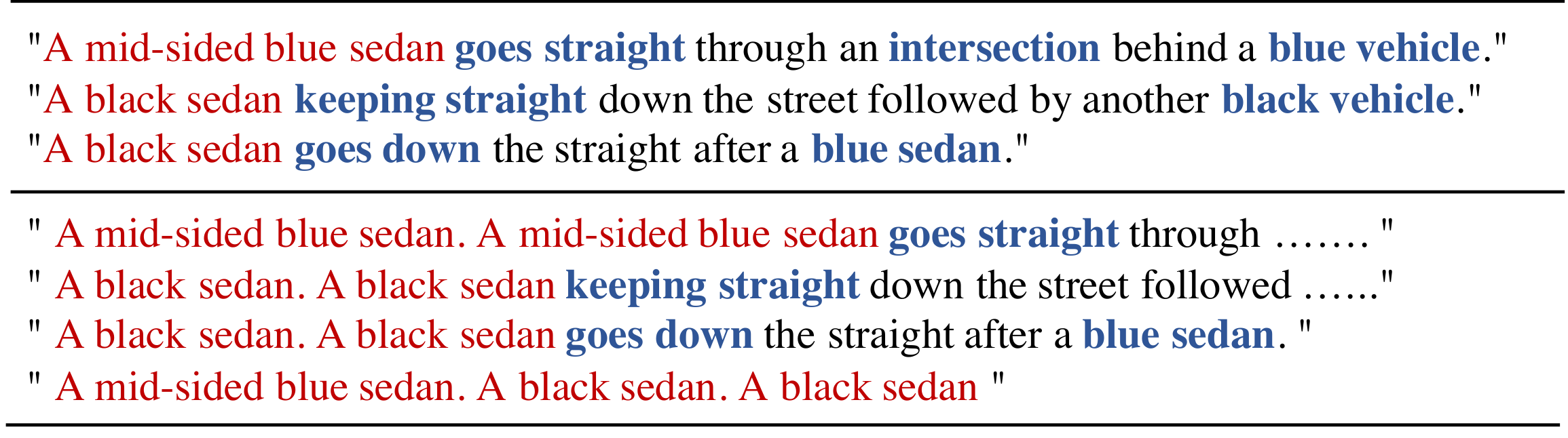}
    \caption{An example of strengthening the subject in the text description.}
    \label{fig:aug_example2}
\end{figure}
In addition, to avoid the interference caused by multiple vehicle descriptions in the text, we enhance the text by strengthening the subject. As shown in Figure~\ref{fig:aug_example2}, the target vehicle is often the subject of the text description. Therefore, we use "spacy "\footnote{\tiny\url{https://spacy.io/}} extract the subject as a separate sentence and put it at the beginning. At the same time, the subject of the three sentences forms the fourth text description.

\subsection{Cross-Modal Representation Learning}

Natural language-based vehicle retrieval aims to retrieve the specific vehicle according to the text description. These texts describe the inherent attributes of the vehicles (\eg, color, type, and size), as well as external factors such as the behavior of the vehicle and the surrounding environment. At the network structure level, we construct a dual-stream image encoder for visual representation learning, the dual-stream architecture takes local detected objectives, \ie, vehicle, as input and global motion images separately to pay attention to the inherent attributes and external factors of the vehicles. In addition, the pretrained text encoder is utilized to extract text embedding. We revisit several losses in terms of language-based vehicle retrieval. For instance, following the natural language supervision in CLIP \cite{radford2021learning}, the symmetric InfoNCE~\cite{oord2018representation} loss is adopted to learn multi-modal embedding space by jointly training an image encoder and text encoder to maximize the cosine similarity of the image and text embeddings. Furthermore, we introduce the instance loss~\cite{zheng2017dual} to learn the instance-level features and the self-supervised barlow twins loss~\cite{zbontar2021barlow} to demand the feature to be as diverse as possible.

\noindent\textbf{Network structure.} 
Following existing re-id works, we adopt the strong  networks pretrained on ImageNet~\cite{deng2009imagenet} as the vision backbone module. We have adopted two vision structures: 
\begin{itemize}
    \item The vanilla re-id baseline\footnote{\tiny\url{https://github.com/layumi/Person_reID_baseline_pytorch}} is used to extract the visual representation for cropped vehicle. We follow the code\footnote{\tiny\url{https://github.com/layumi/AICIty-reID-2020}} of the 4\emph{th} AICity vehicle re-id to pre-train the model on the data of track2 this year. In particular, the visual backbone is SE-ResNeXt50 \cite{hu2018squeeze}. We then fine-tune the model on the track5 data with instance loss to extract the visual feature. Only the cropped vehicle images are considered in this baseline model. We extract the final 512-dim feature before the classification layer as the visual representation. 
    \item To learn the motion information, we further adopt a dual-stream structure. The inputs consist of local cropped images and global motion images. The local cropped images are the detected vehicles cropped from a random frame. The global motion images are generated as illustrated in Section 3.1.1. The dual-stream structure contains two independent CNN encoders pretrained on ImageNet as the backbone, including SE-ResNeXt50 \cite{hu2018squeeze} or EfficientNet B3 \cite{tan2019efficientnet}. In particular, for each stream, we introduce projection heads to map visual representation to the spaces of contrastive representation learning and instance fine-grained feature learning. The projection head uses a MLP with the hidden layer to obtain
\begin{equation}
 z_i = g_i(h_i) = W_2\sigma(BN(W_2 h_i)), \label{eq:image}
 \end{equation}
 where BN is a Batch Normalization (BN) layer, $\sigma$ is a ReLU layer, and the output dimension is 512. $h_i$ is the visual features extracted by the backbone. As shown in Figure \ref{fig:framework}, there are three projection heads, corresponding to local detail features, global motion features and fusion features. In addition, the classification heads are applied to output the predicted possibility of different tracks. The classification head is similar with the projection head, but the output dimension is the number of tracks.
 \end{itemize}
  
 For text embeddings, we deploy pretrained BERT \cite{devlin2018bert} or RoBERTa \cite{liu2019roberta} as text encoder. Similar with the image encoder, the projection head is introduced to map text embeddings to the space of contrastive representation learning. But the BN is replaced with the Layer Normalization (LN) layer. Due to the limited amount of text data, the parameters of text encoder are fixed or updated with a small learning rate.
 \begin{equation}
 z_t = g_t(h_t) = W_2\sigma(LN(W_2 h_t)), \label{eq:text}
 \end{equation}
 where $h_t$ is the text embeddings extracted by the pretrained model.
 
\subsection{Optimization Objectives}
\subsubsection{Contrastive Loss}
To maximize the cosine similarity of the image and text embeddings, we utilize symmetric InfoNCE \cite{oord2018representation} loss like CLIP \cite{radford2021learning}. Specially, we optimize the symmetric InfoNCE in three levels to achieve well-aligned with the given description: (1) local cropped image region and sentence, (2) global motion image and sentence, (3) fusion feature and sentence.
We define the score function following previous work in contrastive learning:
 \begin{equation}
\mathcal{S}  = cos(z_{img},z_{text})/\tau,
 \end{equation}
where $cos(u,v) = \frac{u^Tv}{||u|| ||v||}$ denotes cosine similarity, and $\tau$ denotes a temperature learnable parameter initialized with 1. This maps the image and text representations into a joint embedding space.

Given a batch of $M$ image-text pairs, it consists of $M \times M$ possible sample pairs. The symmetric InfoNCE has two parts: Text-to-Image and Image-to-Text. Text-to-Image compares one positive pair with $M-1$ Negative pairs for each text description:
 \begin{equation}
\mathcal{L}_{t2i}  =\frac{1}{M} \sum_{i=1}^M -\log \frac{exp(cos(z_{img,i},z_{text,i})/\tau)}{\sum_{j=1}^M exp(cos(z_{img,j},z_{text,i})/\tau)}.
 \end{equation}
Meanwhile, Image-to-Text optimizes one positive pair with $M-1$ Negative pairs for each track:
 \begin{equation}
 \mathcal{L}_{i2t}  =\frac{1}{M} \sum_{i=1}^M -\log \frac{exp(cos(z_{img,i},z_{text,i})/\tau)}{\sum_{j=1}^M exp(cos(z_{img,i},z_{text,j})/\tau)}.
 \end{equation}
 The symmetric InfoNCE is formulated as:
  \begin{equation}
  \mathcal{L}_{SNCE}  = \lambda_1 \mathcal{L}_{t2i}+\lambda_2 \mathcal{L}_{i2t},
    \label{nce}
   \end{equation}
   where $\lambda_1, \lambda_2$ are the weights of Text-to-Image and Image-to-Text. Due to the evaluation of Text-to-Image manner, we set $\lambda_1=2, \lambda_2=1$.
   
\noindent\textbf{InfoNCE loss between local cropped image and sentence.} The local cropped image contains the inherent attributes of the vehicle (e.g., color, type, and size). These inherent attributes should be consistent with corresponding words in the text description, which are often the text description about the subject. In order to strengthen the relationship between the target vehicle image and the subject of the text description, we adopt the description augmentation in Section 3.1.2.

\noindent\textbf{InfoNCE loss between global motion image and sentence.} The global motion image reflects the motion of vehicle and the external factors of the surrounding environment. As illustrated in section 3.1.1, our motion images can effectively retain these information, and the role of these external factors can be grasped through the supervision of the global motion map. .
   
\noindent\textbf{InfoNCE loss between fusion feature and sentence.} Fusion of local and global features using nonlinear mapping takes advantage of neural networks to mine some complex associations. During the inference, we only use the fused features as the representation of the retrieval.
   
   At three levels, we use the symmetric InfoNCE in Eq. \ref{nce} to optimize the learning of contrastive representations, and the weight of each level is 1.
   
\subsubsection{Instance Loss}
Instance loss is one common objective in the bi-directional image-text retrieval task to capture the global discrepancy ~\cite{zheng2017dual}, and we also explore this loss in terms of the natural language-based vehicle retrieval task. We treat every track and the corresponding descriptions as one category. The optimization goal is to mapping the visual and textual input into one shared classification space. In particular, we adopted one shared classifier for both visual and textual inputs, and enforce the model to learn the mapping function.
\begin{align}
      \mathcal{L}_{i} = -log(W_{shared}~z_{i}), \\
    \mathcal{L}_{t} = -log(W_{shared}~z_{t}),  
\end{align}
where $z_{i}$ and $z_{t}$ are the visual and textual embedding defined in Eq.~\ref{eq:image} and Eq.~\ref{eq:text}, and $W_{shared}$ denotes the weight of the final linear classifier. Instance loss can be formulated as:
\begin{equation}
    \mathcal{L}_{instance} = \mathcal{L}_{i} + \mathcal{L}_{t}.
\end{equation}
It is worth noting that the instance loss is different from the symmetric infoNCE in whether it optimizes the cosine similarity within one mini-batch or the stored classification weights of all image-text pairs. As shown in Table~\ref{table:ab}, the instance loss is complementary to the contrastive loss, which further boosts the performance. 


\subsubsection{Barlow-twins Loss}
Barlow-twins loss~\cite{zbontar2021barlow} is an optional loss in the proposed framework. We trained three models based on such loss for the ensemble. This loss is similar to the CLIP loss~\cite{radford2021learning} but it conducts the feature multiplication in the feature channel, which results in one totally different request to the learned feature. Actually, this loss can be viewed as one regularization term. It asks the model to learn one orthogonal feature, where every channel contains a different semantic meaning from the rest channels. 

\section{Experiment}
\subsection{Dataset Analysis}

Natural language (NL) description offers another useful way to specify vehicle track queries. The dataset for Natural Language-Based Vehicle Retrieval Track is built upon the CityFlow Benchmark by annotating vehicles with natural language descriptions. This dataset contains 2498 tracks of vehicles with three unique natural language descriptions each. 530 unique vehicle tracks together with 530 query sets with three descriptions are curated for this challenge.

\begin{figure}[tb]
    \centering
    \includegraphics[width=1.0\linewidth]{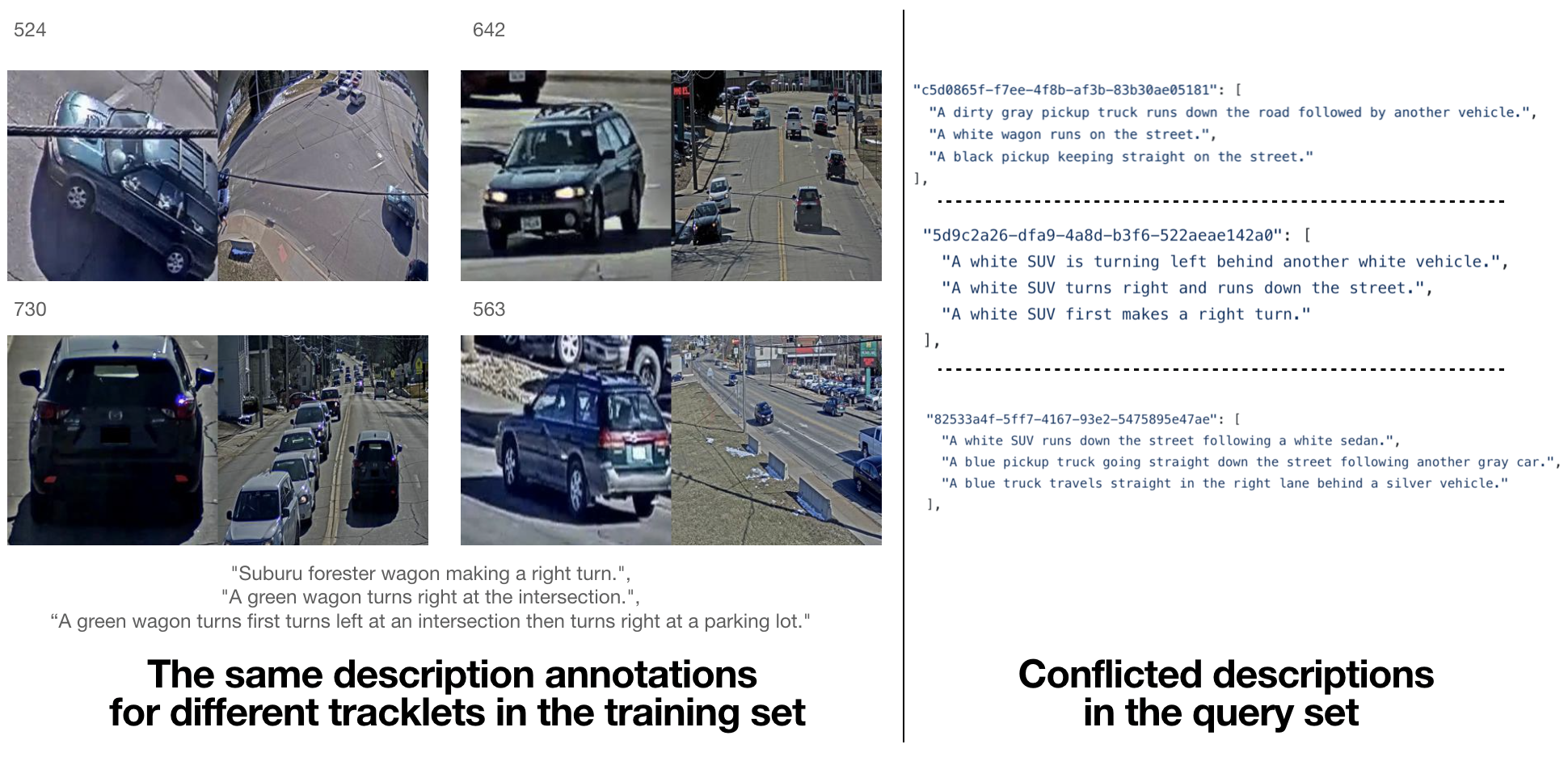}
    \caption{Noise in the training \& test set. We observe two kinds of noise existed in the dataset. The identical three descriptions are annotated for different vehicle tracklets in the training set (left). Conflicting descriptions existed in the query set (right). }
    \label{fig:noise}
\end{figure}

\noindent\textbf{Noise in the training \& test set.} As shown in Figure~\ref{fig:noise}, we observe that noise exists in both training and test set. The main noise is from the same three descriptions for different vehicle tracklets. There are $323$ tracklets sharing the identical three sentences with another tracklets. A similar phenomenon is observed on the query set, and there are $56$ queries containing the identical three textual descriptions. Such textual input compromises the training process as well as the inference accuracy. Besides,  we also observe the conflicting descriptions in the query set. For instance, ``turning left'' and ``turns right''  simultaneously appear in one description group. We can not optimize such noise but deploy the mean textual feature to find the most similar samples. 

\noindent\textbf{Evaluation.} The Vehicle Retrieval by NL Descriptions task is evaluated using standard metrics for retrieval tasks. The Mean Reciprocal Rank (MRR) is used \cite{voorhees1999trec}. It is formulated as :
 \begin{equation}
MRR=\frac{1}{|Q|} \sum _{i=1}^{|Q|}\frac {1}{rank_{i}},
 \end{equation}
where $rank_{i}$ refers to the rank position of the right track for the $i_{th}$ text description, and $Q$ is the set of text descriptions.

\begin{table}[t]
\scalebox{1}{
\begin{tabularx}{8.5cm}{p{1.cm}<{\centering}|X<{\centering} |X<{\centering}}
\toprule[1.5pt] 
Rank&Team Name &MRR\\
\midrule[1pt]
1&  \bf{Alibaba-UTS (Ours)}&  \bf{0.1869}\\
2&  TimeLab&  0.1613\\
3&  SBUK&  0.1594\\
4&  SNLP&  0.1571\\
5&  HUST&  0.1564\\
\bottomrule[1.5pt]
\end{tabularx}}
  \caption{Competition results of AI City Natural Language-Based Vehicle Retrieval Challenge.}
   \label{result}
\end{table}

\begin{table}[t]
\setlength{\belowcaptionskip}{2pt}
\scalebox{1}{
\begin{tabularx}{8.5cm}{p{3.cm}<{\centering}|X}
\toprule[1.5pt] 
Method& Performance\\
\midrule[1pt]
Baseline&   \checkmark \ \quad \checkmark \ \quad \checkmark \ \quad \checkmark \ \ \quad \checkmark \ \quad \checkmark\\
Instance Loss &  \quad  \quad \ \checkmark \ \quad \checkmark \ \quad \checkmark \ \quad \checkmark \ \quad \checkmark \\
Motion Image&   \quad  \quad \quad \quad \  \checkmark \ \quad \checkmark \ \ \quad \checkmark \ \quad \checkmark \\
NLP Augmentation&    \quad  \quad \quad \quad \quad \quad \ \checkmark \ \ \quad  \checkmark \ \quad \checkmark \\
Large Size\& Model&   \quad  \quad \quad \quad \quad \quad \quad \quad \ \  \checkmark \ \quad \checkmark\\
Ensemble &   \quad  \quad \quad \quad \quad \quad \quad \quad \quad  \ \  \quad \checkmark\\
\midrule[1pt]
MRR(\%) &   \small{8.25} 9.65 13.21 14.56 19.27 20.77\\
\bottomrule[1.5pt]
\end{tabularx}}
  \caption{Ablation Study on TestA in the online evaluation system.}
   \label{table:ab}
\end{table}

\begin{figure*}[t]
\begin{center}
\includegraphics[width= 17cm]{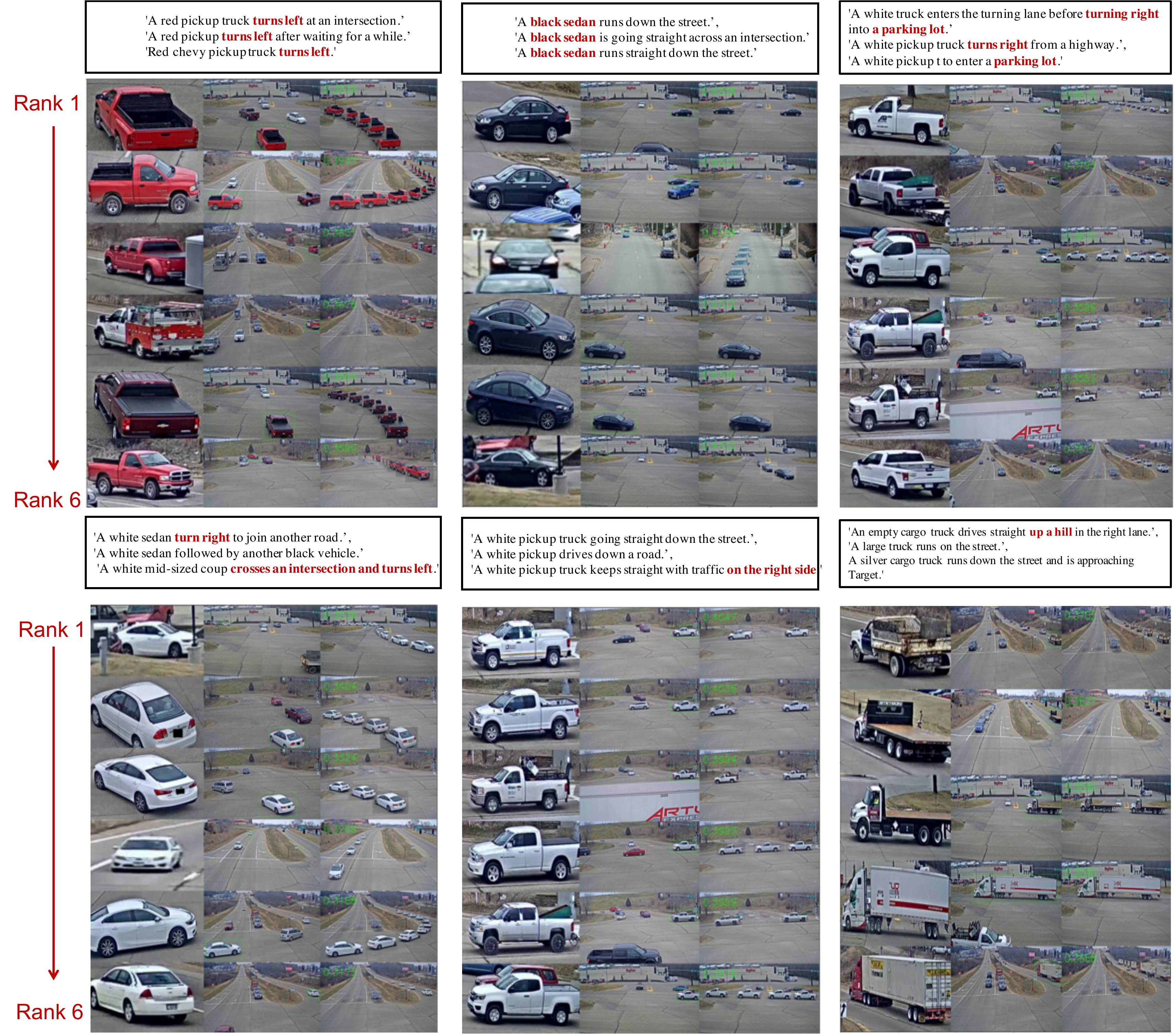}
   \caption{\textbf{Qualitative Results.} We highlight ``key'' words in \textcolor{red}{red} and show the proposed method can find the right matches with the fine-grained attention.}
   \label{fig:vis}
\end{center}
\end{figure*}

\subsection{Quantitative Results}

During the inference, we average all the frame features of the target in each track as track features, the embedding described by three sentences is also averaged as the query features. The cosine distance is used for ranking as the final result.

\noindent\textbf{Comparsion with Other Teams.} As shown in Table \ref{result}, the proposed method has achieved the state-of-the-art MRR, i.e., 0.1869, which is superior to the second-best team by a large margin. Moreover, the consistent performance on all Test datasets demonstrates the effectiveness and robustness of the proposed method.

\noindent\textbf{Abation Study.}  As illustrated in Table \ref{table:ab}, we perform ablation studies with different modules of our proposed method. The ``Baseline" donates that CLIP \cite{radford2021learning} with ResNet50 as image encoder and BERT-BASE as text encoder. The ``Instance Loss" optimizes the cross-entropy loss function for distinguishing each track.``Motion Images" donates the dual-stream architecture with local cropped image and global motion images. The introduction of motion images gains a relative MRR improvement of 36.5\%, which demonstrates the external factors and motion information play a vital role in natural language-based vehicle retrieval. Our method of motion and background modeling is a simple and effective manner to capture these information. The ``NLP augmentation" consists of strengthening the subject description and backtranslation. ``Large Size\&Model" means that using larger pretrained models, such as RoBERTa \cite{liu2019roberta} as text encoder and ResNeXt101 as image encoder. The size of image input is improved to $320 \times 320$. The obvious improvement of larger pretrained model proves the importance of Large-scale language pre-training. The well-pretrained text encoder provides a good Initialization embedding space, especially in the case of lack of text content. On the 50\% Test set, we improve the CLIP baseline from 0.0825 to 0.1927 mAP MRR with single model.

\subsection{Qualitative Results}
Furthermore, we visualize the ranking results in Figure~\ref{fig:vis}, which shows the effectiveness of the proposed method. All the top-6 samples are relevant to the query descriptions. We highlight ``key'' words in the description and show the proposed method can find the right matches with the fine-grained attention.

\section{Conclusion}
In this paper, we propose a robust natural language-based vehicle search system for smart city applications. To connect the vision and language modalities, we jointly train the state-of-the-art vision model and transformer-based language model with the symmetric InfoNCE loss and instance loss. Further, we design a two-stream architecture to incorporate both local details and global information of vehicles, and apply the text augmentation technique backtranslation to enhance the model robustness.
Finally, the system achieves 18.69\% MRR accuracy and reaches the first place in the natural language-based vehicle retrieval track of the 5th AICity Challenge.
In the future, we will continually explore the large-scale and efficient vehicle search technique for the intelligent transportation system, such as more elaborate model architectures, more powerful optimization objectives and more abundant data augmentation methods.

{\small
\bibliographystyle{ieee_fullname}
\bibliography{egbib}
}

\end{document}